# Evaluating Human and Machine Confidence in Phishing Email Detection: A Comparative Study

Paras Jain*
*Rochester Institute of Technology*
USA
pj2196@rit.edu

Khushi Dhar*
*Rochester Institute of Technology*
USA
kd4849@rit.edu

Olyemi E. Amujo*
*Rochester Institute of Technology*
Rochester, NY, USA
oa6121@rit.edu

Esa M. Rantanen
*Rochester Institute of Technology*
USA
esa.rantanen@rit.edu

*Abstract*—Identifying deceptive content like phishing emails demands sophisticated cognitive processes that combine pattern recognition, confidence assessment, and contextual analysis. This research examines how human cognition and machine learning models work together to distinguish phishing emails from legitimate ones. We employed three interpretable algorithms Logistic Regression, Decision Trees, and Random Forests training them on both TF-IDF features and semantic embeddings, then compared their predictions against human evaluations that captured confidence ratings and linguistic observations. Our results show that machine learning models provide good accuracy rates, but their confidence levels vary significantly. Human evaluators, on the other hand, use a greater variety of language signs and retain more consistent confidence. We also found that while language proficiency has minimal effect on detection performance, aging does. These findings offer helpful direction for creating transparent AI systems that complement human cognitive functions, ultimately improving human-AI cooperation in challenging content analysis tasks.

## I. INTRODUCTION

Identifying deceptive or misleading email communication requires sophisticated cognitive abilities such as pattern recognition, contextual inference, and confidence assessment. Machine learning (ML) models have shown promise in automating such tasks, yet understanding how these models' processes align with human reasoning remains a crucial research frontier. Investigating this alignment not only improves model transparency and interpretability but also enhances human-machine collaboration in decision-making.

This research was supported by the National Science Foundation under Award No. DGE-2125362. Any opinions, findings, and conclusions or recommendations expressed in this paper are those of the authors and do not necessarily reflect the views of the National Science Foundation.
*Paras Jain, Khushi Dhar, and Olyemi E. Amujo contributed equally to this work.

### A. Purpose of the Research

This study examined the cognitive mechanisms underlying phishing email detection by comparing human annotations with predictions from interpretable machine learning models. We focused on three key cognitive aspects: the linguistic features that guide decision-making, confidence calibration in judgments, and demographic influences such as age and language background. By analyzing both machine and human reasoning strategies, we aimed to identify gaps and overlaps in human and machine resoning that can inform the development of AI systems that resonate more closely with human cognition.

Our approach involved training Logistic Regression, Decision Tree, and Random Forest classifiers on textual data represented via Term Frequency-Inverse Document Frequency (TF-IDF) and sentence embeddings. Alongside model predictions, human participants provided annotations including confidence ratings and linguistic cues, enabling a multifaceted comparison. The findings reveal that while models excel in classification accuracy, humans maintain steadier confidence and draw on more diverse linguistic information than the ML models, highlighting complementary strengths of both.

### B. Research Questions

We address the following research questions:
1) How do human and machine cognition compare in processing linguistic cues for deception detection?
2) How do confidence levels differ between humans and machines when making these judgments?
3) What cognitive effects do age and language background have on detection accuracy?
4) How can understanding these differences enhance human-machine collaboration?

By focusing on cognition and behavioral insights, this work contributes to the broader understanding of AI as a partner in complex interpretive tasks, paving the way for more transparent, trustworthy, and human-centered intelligent systems.

## II. REVIEW OF RELEVANT LITERATURE

Phishing attacks exploit human cognitive vulnerabilities, often relying on deceptive language and psychological manipulations to extract sensitive information. Traditional spam detection systems, which largely depended on keyword matching and rule-based filters, have proven insufficient against increasingly sophisticated phishing tactics. Uplenchwar et al. [1] developed a phishing detection system (PADSTM) using machine learning techniques such as Random Forests, Naive Bayes, Support Vector Classification, and K Nearest Neighbor, demonstrating high performance (up to 97% accuracy) through customized feature extraction including tokens, URLs, and currency symbols. However, limitations like keyword dependency and the lack of contextual understanding highlight gaps in addressing phishing that leverages subtle human psychological cues.

Advancements in AI, particularly transformer-based language models like BERT, offer promising alternatives by capturing linguistic nuances and semantic context. Rifat et al. [2] explored BERT and DistilBERT for SMS phishing detection, achieving 98.21% accuracy and outperforming traditional models. The study emphasized using an imbalanced dataset to reflect real-world scenarios but acknowledged the challenges in generalizability and the absence of interpretability, especially important in cybersecurity. Despite these advances, such models remain largely black-boxes in nature, leaving questions about how their decision-making aligns with or diverges from— human cognitive processes.

Understanding how AI compares to human cognition is essential, especially as phishing relies on exploiting specific cognitive biases. Rodríguez-Priego et al. [3] showed that loss-framed messages could better influence cybersecurity behavior, reinforcing how message framing impacts human decisions. Flusberg & Holmes [4] tested large language models against human linguistic framing effects, finding partial alignment, while Berto & Özgün [5] used logic-based frameworks to explain why humans respond differently to logically equivalent information. These studies underscore that while AI can mimic some cognitive patterns, it lacks deeper human-like contextual reasoning. Therefore, comparing AI detection mechanisms to human susceptibility remains a vital research direction in improving phishing defense systems.

Beyond algorithmic detection, understanding human susceptibility to phishing is critical. Several studies have examined demographic and cognitive factors that shape phishing awareness and response. Daengsi et al. [6] conducted a large-scale study in Thailand's financial sector and found that women showed higher post-training awareness than men, while age differences diminished after education, suggesting the power of targeted interventions. Similarly, Grilli et al. [7] used the Phishing Email Suspicion Test (PEST) to reveal that older adults had lower discrimination ability despite similar bias levels across ages, highlighting age-related cognitive limitations.

Language and culture also play essential roles in phishing detection, particularly for non-native English speakers (NNESs). Hasegawa et al. (2021) [8] investigated how non-native English speakers (Germans, Japanese, and Koreans) respond to phishing emails in both English and their native languages. Their survey of 862 participants revealed that Japanese users were more risk-prone when reading English emails, whereas Korean users tended to be less risk-prone in that context. Their findings showed that linguistic context influenced cue reliance, with Japanese users becoming more risk-prone in English, while Koreans became less so in that context. This highlights how cultural and linguistic backgrounds significantly influence reliance on phishing cues and suggests that one-size-fits-all indicators may not be universally effective. Complementing this, An et al. [9] demonstrated that combining OSINT metadata with Random Forest and XGBoost substantially improves detection across both English and Arabic text.

Uplenchwar et al [10] proposed the PADSTM system, which combines keyword-based heuristics, URL blacklists, and Random Forest/SVM classifiers to detect phishing attacks with 97% accuracy, but its dependence on fixed features may reduce adaptability against new threats [10]. More recently, Somesha & Pais presented DeepEPhishNet, leveraging Word2Vec and FastText embeddings with ensemble deep models (DNN and BiLSTM) to achieve nearly 99% accuracy on email datasets. However, they also reported elevated false-positive rates and noted the need for testing in real-world deployment scenarios [11].

While prior work has either emphasized high-performing but opaque models—such as BERT-based architectures for phishing detection [2]—or investigated cognitive and demographic factors influencing phishing susceptibility [12], [13], few studies have directly juxtaposed human judgments with transparent, interpretable machine learning models on the same dataset. To address this gap, our methodology integrates interpretable classifiers (Logistic Regression, Decision Tree, and Random Forest) with fine-grained human annotations that capture not only classification labels but also confidence levels, emotional responses, and self-identified linguistic cues.

This side-by-side evaluation enables a nuanced comparison of human and machine decision-making, revealing not just performance differences but the underlying reasons for divergence. By analyzing the decision features, emotional triggers, and rationale used by both humans and models, we provide a multifaceted view of phishing detection behavior. This dual-lens design contributes actionable insights toward (i) developing more human-aligned explainable AI systems, and (ii) designing tailored user education interventions that emphasize the cues commonly missed by users. In doing so, our work advances both computational models of phishing detection and human-centered defense strategies.

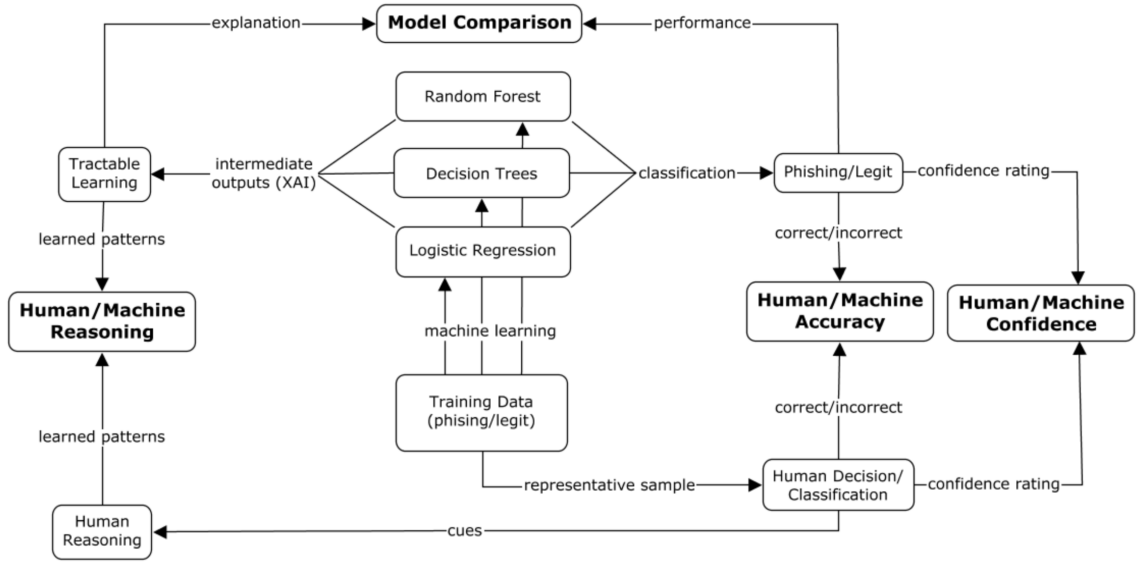

Fig. 1. Overview of the research plan, with multiple comparisons between humans and ML models as well as between three specific ML models.

## III. METHODOLOGY

Our Methodology is an amalgamation of machine learning methods that are transparent so that we can have understanding of the tractable learning process for detecting phishing and non-phishing emails and compare it to that of a human. We compared Random Forest, Decision Trees, and Logistic Regression models as well as human reasoning, accuracy, and confidence to these models.

### A. Data Collection

For model training and evaluation we used the Enron email dataset [14], which includes email content along with binary labels indicating phishing or legitimate emails. To understand human perspective, we generated 10 phishing and 10 non-phishing emails using ChatGPT. **In both the datasets, the phishing emails were depicted by 1 and non-phishing with 0**. These emails were given as a questionnaire to the users to perform various annotations. These annotations captured human-predicted label (phishing or not phishing), confidence in their annotation (0-100%), linguistic cues within the email that prompted the particular annotation, and emotion invoked by the email. Both human participants and machine learning models were evaluated on the exact same set of emails. This one-to-one correspondence ensured that any observed differences in performance, confidence, or linguistic cue usage were attributable to differences in reasoning processes rather than differences in the input data. Such alignment provides a controlled basis for comparing cognitive and algorithmic approaches.

### B. Data Preprocessing

All e-mail bodies were pre-processed to reduce lexical sparsity while preserving content indicative of phishing. Pre-processing was implemented in Python using the re, nltk, and BeautifulSoup libraries. The steps were as follows:

1) E-mail and URL removal: Regular expressions were used to remove all email addresses and URLs.
2) HTML tag removal: HTML content was stripped using `BeautifulSoup`.
3) Normalization: Text was converted to lowercase; punctuation and numeric tokens were removed.
4) Tokenization and lemmatization: Words were tokenized and lemmatized using `nltk.WordNetLemmatizer` was used to reduce inflectional variations and unify word forms.

This preprocessing pipeline ensures consistency across the dataset while retaining meaningful features for phishing detection.

### C. Feature Extraction

Two representations of email text were generated for each of the three machine learning models:

1) **Term Frequency–Inverse Document Frequency (TF-IDF) Vectors**
   - Implemented with `sklearn.feature_extraction.text.TfidfVectorizer`.
   - Parameters:
   (a) `max_features=5000`
   (b) `ngram_range=(1,2)` to capture unigram and bigram.
   (c) `lowercase=False` since we lowercased earlier.

The vectorizer was applied to pre-lowercased, cleaned text, converting each email into a sparse numerical vector. This approach emphasized terms that are frequent in individual emails but rare across the dataset, improving discrimination between phishing and non-phishing content across the corpus.

2) **Sentence-BERT Embeddings**
   - Implemented using `sentence-transformers` library with the pretrained model `'all-MiniLM-L6-v2'`.
   - Each email was encoded into a 384-dimensional dense vector.
   - Embeddings were normalized to unit length to ensure comparability in vector space using `normalize_embeddings=True`.

These representations captured both lexical frequency patterns (TF-IDF) and semantic context (S-BERT) for downstream classification.

### D. Data Splitting

The dataset was split into training and validation sets using a 70:30 ratio with a fixed `random state = 42` for reproducibility.

### E. Model Training

We employed three supervised learning models to classify emails as phishing or safe: Logistic Regression, Decision Tree, and Random Forest. Each model was trained on TF-IDF representations and semantic embeddings. Below, we describe the configuration and rationale for each model in detail.

*1) Logistic Regression:* Logistic Regression is a well-established linear classification algorithm that models the relationship between input features and class labels using a logistic (sigmoid) function. It is particularly effective for binary classification tasks, making it suitable for distinguishing phishing from legitimate emails.

The Logistic Regression model was configured as follows:
- `max_iter=1000` to ensure convergence during training.
- `class_weight='balanced'` to address the imbalance between phishing and safe emails.
- `C=1.0` for L2 regularization strength to prevent overfitting by penalizing large weights in the model.

*2) Decision Tree Classifier:* Decision Tree is a non-linear, tree-based model that recursively partitions the feature space using feature values to classify samples. It creates branches and leaf nodes corresponding to decision rules learned from the training data.

The Decision Tree model was configured as follows:
- `sklearn.tree.DecisionTreeClassifier` was used for implementation of the tree.
- The random state was set to 42 to ensure reproducibility.
- We used `gini` as default to measure the quality of a split.
- All other parameters (`max_depth`, `min_samples_split`, etc.) were left at scikit-learn defaults.

*3) Random Forest Classifier:* Random Forest is an ensemble learning method that constructs multiple decision trees and aggregates their predictions to improve classification accuracy and reduce overfitting. Each tree is trained on a bootstrap sample of the data and considers a random subset of features when splitting nodes.

The Random Forest model was configured as follows:
- Implementation was done through `sklearn.ensemble.RandomForestClassifier`.
- `n_estimators=100` was used to set the number of trees to be trained at 100.
- Random state was set to 42 to ensure reproducibility.
- All other parameters (`max_depth`, `min_samples_split`, etc.) were left at scikit-learn defaults.

The trained models was saved using `joblib.dump()`, and the Sentence-BERT embedder was saved separately using `embedder.save()`.

### F. Model Evaluation

Each model was evaluated on both the training and validation sets. While accuracy measures overall correctness, F1 score provides a more balanced evaluation of precision and recall, which is crucial in phishing detection where false positives and false negatives have significant implications. Therefore, F1-score is the preferred metric for evaluating the classification performance in this study. These metrics offer insights into each model's performance, particularly their ability to correctly identify phishing emails without excessive false positives.

The models were then evaluated on a separate test set which was generated from ChatGPT consisting of 10 phishing and 10 non-phishing emails. The email bodies were preprocessed and transformed using the same TF-IDF vectorizer applied during training. Predictions and class probabilities were obtained for the test emails, allowing assessment of model performance on unseen data and saved in the form of pickle files.

On the other hand, human performance was measured on the test data using majority vote labels for each email, which were then directly compared with model predictions on the same inputs. This parallel evaluation method eliminated bias induced by the data set and allowed a clearer interpretation of reasoning differences.

### G. Confidence Scoring and Feature Importance Analysis

To extract confidence scores, the `predict_proba()` function was applied. It returned the probability estimates for each class, allowing us to measure the model's certainty in its predictions. These scores were scaled to a percentage format (0–100).

For the TF-IDF-based models, the most influential words in each email were determined by selecting the features with the highest importance scores among the terms present in the email, based on the model's `feature_importances_`. These per-email important features were extracted and stored for later comparison with human annotations.

For embedding-based models, per-email important words were not computed since each TF-IDF dimension corresponds to a specific word or n-gram, allowing us to calculate contributions to a prediction using model coefficients or feature importances. In contrast, embeddings produce dense, abstract vectors where each dimension encodes semantic information across the entire text, making it impossible to directly attribute importance to individual words.

## IV. RESULTS

Table I presents the F1-scores achieved by the Logistic Regression (LR), Decision Tree (DT), and Random Forest (RF) classifiers across both phishing and non-phishing classes, using two feature extraction approaches: TF-IDF and sentence-level embeddings.

TABLE I
F1-SCORES FOR PHISHING AND NOT PHISHING CLASSES BY MODEL AND FEATURE TYPE

| Model | F1-score (Phishing) | F1-score (Not Phishing) |
|---|---|---|
| LR with TF-IDF | 0.72 | 0.53 |
| LR with Embeddings | 0.67 | 0.50 |
| DT with TF-IDF | 0.73 | 0.67 |
| DT with Embeddings | 0.59 | 0.70 |
| RF with TF-IDF | 0.73 | 0.67 |
| RF with Embeddings | 0.70 | 0.70 |

From Table I, we observe that models trained using TF-IDF features generally outperform their embedding-based counterparts in phishing detection, particularly for the Logistic Regression and Decision Tree models. Logistic Regression with TF-IDF achieved an F1-score of 0.72 for phishing emails, compared to 0.67 with sentence embeddings. This trend suggests that TF-IDF, despite being a simpler lexical representation, retains important discriminative signals for phishing-related vocabulary.

The Decision Tree and Random Forest models exhibited higher performance than Logistic Regression in the phishing class. Notably, Random Forest with embeddings achieved balanced performance, with an F1-score of 0.70 in both classes, indicating it effectively captured both phishing and legitimate patterns in the data.

Overall, Random Forest and Decision Tree models with TF-IDF and embedding features, respectively, demonstrated the most robust and interpretable results, confirming the effectiveness of tree-based classifiers for phishing email detection when paired with sparse lexical representations.

In subsequent analysis, we compared these models' outputs against human annotators.

To evaluate human performance, we computed majority-vote labels across participants for each email and measured their agreement with ground truth as can be seen in Figure 3.

We then compared the prediction accuracy between the machine learning models (with TF-IDF). As can be seen in

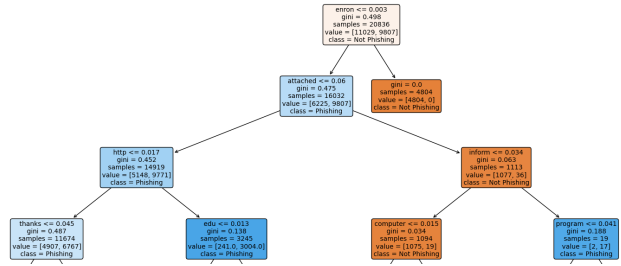

Fig. 2. First 3 branches of prediction in Decision Tree

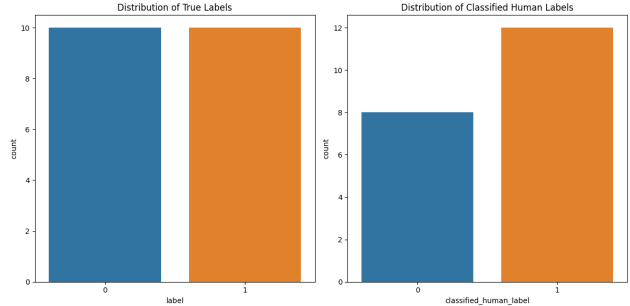

Fig. 3. True vs human label predictions (0 = Non-Phishing, 1 = Phishing)

Figure 3, Logistic Regression predicted the most non-phishing while humans performed equivalent to Decision Trees and Random Forest.

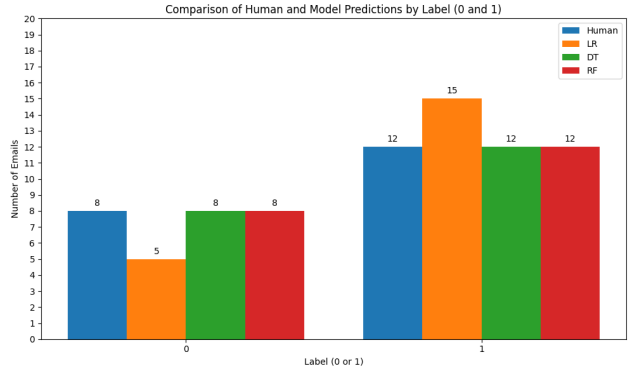

Fig. 4. ML Models vs Human Label predictions

In analyses of the confidence of both subjects (machines and humans), it was quite interesting to find that even though the tree based models had the same performance as humans, they lacked confidence in their detection while humans showed consistent confidence in 60-80% on average.

Additionally, the important words which support the human or model's decision making were also analyzed.

Humans showed more diversity in terms of word importance when it came to differentiating between phishing and non-phishing emails.

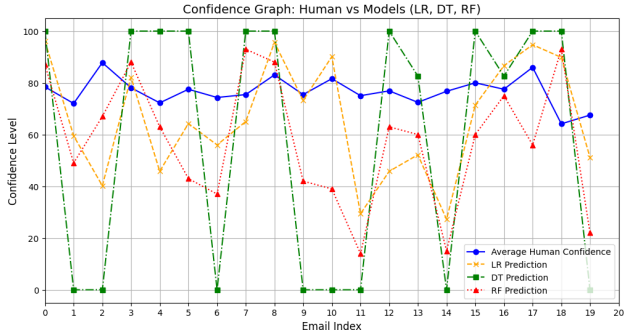

Fig. 5. ML Models vs Human Label confidence score

TABLE II
COMPARISON OF MOST COMMON WORDS SELECTED BY MODEL VS.
HUMANS FOR PHISHING EMAILS

| Machine Word | Frequency | Human Word | Frequency |
|---|---|---|---|
| account | 3 | email | 17 |
| desk | 3 | payment | 12 |
| account | 3 | verify | 17 |
| bank | 3 | update | 22 |
| support | 3 | help | 14 |
| account | 4 | hours | 20 |
| revised | 3 | could | 16 |
| request | 3 | hours | 22 |
| account | 3 | funds | 23 |
| life | 3 | link | 19 |

TABLE III
COMPARISON OF MOST COMMON WORDS SELECTED BY MODEL VS.
HUMANS FOR NON-PHISHING EMAILS

| Machine Word | Frequency | Human Word | Frequency |
|---|---|---|---|
| account | 4 | contact | 19 |
| account | 4 | password | 15 |
| hello | 3 | office | 17 |
| account | 4 | account | 22 |
| schedule | 3 | schedule | 14 |
| attached | 3 | payment | 32 |
| account | 4 | password | 14 |
| company | 3 | payment | 24 |
| attached | 3 | process | 15 |
| employee | 3 | contact | 17 |

In parallel, we explored the impact of age and language in this spectrum.

The highest average accuracy for people was between the ages of 36-45 at 78% followed by 45+ at 74%, 18-25 at 71% and 26-35 at 68%.

TABLE IV
ACCURACY OF PREDICTION BASED ON AGE GROUPS

| Age Group | Accuracy |
|---|---|
| 18–25 | 71% |
| 26–35 | 68% |
| 36–45 | 78% |
| 45+ | 74% |

On the other hand, there was no gap between the accuracy of English speakers and those whose first language was not English. Both were at around 71%. However, it was interesting to notice that English speakers had a higher level of variability which meant that some individuals performed incredibly well having an accuracy individually of 95% as can be seen from the boxplot in Figure 7.

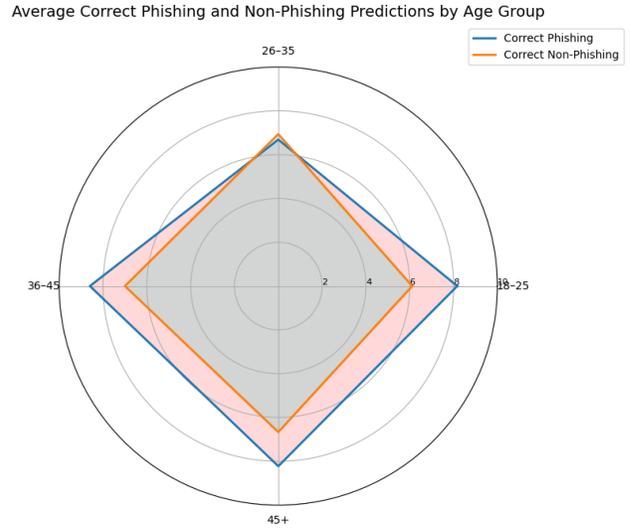

Fig. 6. Correct annotations by different age groups for phishing and non-phishing emails

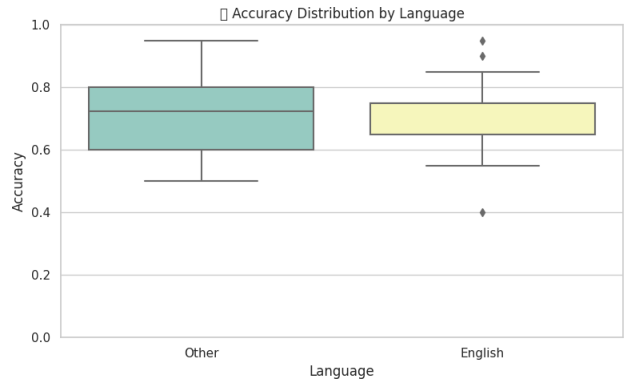

Fig. 7. Correct annotations by different age groups for phishing and non-phishing emails

## V. DISCUSSION

### A. TF-IDF vs Embeddings

The results indicate that models trained with TF-IDF features generally achieve higher F1-scores for detecting phishing emails compared to their embedding-based counterparts, particularly for Logistic Regression and Decision Trees. This suggests that TF-IDF effectively emphasizes the key words and frequency patterns that are most indicative of phishing content. On the other hand, embedding-based models, while slightly less precise for phishing detection, tend to perform better or equally well for the "Not Phishing" class, reflecting their strength in capturing broader semantic context rather than exact keyword frequency. Random Forests show a balanced behavior across both feature types, demonstrating robustness in handling both phishing and non-phishing instances. Overall, these patterns highlight a trade-off: TF-IDF is better at leveraging explicit textual signals that differentiate phishing emails, whereas embeddings provide richer semantic representation, which can improve generalization but may slightly dilute focus on the most discriminative terms.

### B. Human vs. Machine Accuracy

Interestingly, Decision Trees and Random Forests showed performance comparable to human annotators, indicating that these models effectively capture key patterns in the data. However, humans tend to outperform or match machines on certain emails, highlighting the nuanced judgment, contextual understanding, and intuition that humans bring to the task. While machine learning models excel at identifying statistical patterns and frequently occurring cues, they may miss subtleties, such as uncommon phrasing, or context-specific signals that humans can interpret more effectively. This contrast underscores the complementary nature of human and machine performance: machines offer consistency, scalability, and speed across large volumes of data, whereas humans provide critical interpretive skills that help navigate edge cases or ambiguous instances. Overall, the parallel suggests that combining human insight with model predictions could yield more robust and reliable phishing detection systems than relying on either alone.ain emails, reflecting the nuanced judgment humans bring to the task.

### C. Human vs. Machine Confidence

While Logistic Regression achieved the highest phishing detection accuracy and the tree based models showed similar performance to that of humans, they exhibited extremely varied confidence on some predictions compared to human participants, whose confidence levels remained relatively stable between 60-80%. This suggests that human confidence is better calibrated with their actual performance, whereas machine confidence may fluctuate more widely. Such differences highlight the importance of confidence calibration in automated systems to avoid over- or under-trusting model predictions. Consistent human confidence may also reflect the broader linguistic cues and contextual understanding humans use, which machines currently struggle to replicate fully. Improving machine confidence estimation could enhance trust and facilitate more effective human-machine collaboration.

### D. Human vs. Machine Reasoning

A key difference between humans and machines emerged in the diversity of linguistic cues used for phishing detection. Humans demonstrated a broader range of important words when identifying phishing versus non-phishing emails, reflecting richer reasoning strategies that consider multiple contextual signals. In contrast, machine models often rely on a limited set of frequent tokens, which, while effective, may miss subtleties recognized by human annotators. This finding underscores the potential for explainable AI (XAI) techniques and tractable learning approaches to bridge the gap by incorporating interpretable features aligned with human reasoning. Enabling models to explain their decisions transparently can increase user trust and improve decision-making. Future research should focus on integrating diverse linguistic cues and human feedback into model training to develop systems that better emulate human reasoning and confidence calibration.

### E. Human Predictions vs Demographics

Overall, individuals exhibited considerable variation in their ability to discern email types, with some showing a broader range of attention to relevant cues than others. Age appeared to influence prediction accuracy, suggesting that certain older age groups may be more adept at recognizing suspicious content. In contrast, language background did not create a clear advantage, as both native and non-native english speakers performed similarly on average. However, variability within language groups indicated that individuals whose first language was english did better. These differences could lead to strong performance, highlighting that while demographic trends provide some insight, personal experience and attentiveness likely play a critical role in accurate human detection.

### F. Limitations

While our study offers valuable insights by juxtaposing human and machine reasoning in phishing detection, several limitations must be acknowledged.

First, we intentionally focused on interpretable models (Logistic Regression, Decision Trees, and Random Forests) rather than state-of-the-art deep learning architectures. This choice prioritizes transparency and cognitive comparability, allowing us to analyze feature importance and decision pathways. While this may under-represent the predictive accuracy of advanced models, it enables a meaningful comparison between human reasoning and machine decision-making, aligning with our focus on the cognitive aspects of phishing detection.

Second, the human annotation dataset was small and synthetic, comprising 20 emails with balanced phishing and legitimate examples. This design allowed us to control linguistic and structural variables, manage participant workload, and create a standardized framework for comparison. We acknowledge that this dataset does not capture the full complexity, noise, or evolving tactics of real-world phishing campaigns. Future work should extend these analyses to larger, authentic datasets to validate and generalize the findings.

Third, human confidence ratings were self-reported and may include subjective bias, even though we collected demographic information such as age and first language. While these factors help contextualize performance differences, the ratings still reflect individual perception rather than purely objective measures of decision certainty. Future studies could complement self-reports with behavioral or physiological indicators—such as reaction times, hesitation patterns, or eye-tracking—to more accurately capture human decision strategies.

Lastly, although human participants submitted their annotations via an online survey, this controlled setting may not fully reflect real-world email interactions. In practice, users are subject to contextual factors such as multitasking, notifications, interface design, and fatigue, which could influence detection accuracy and confidence. Future work could incorporate in-situ or real-time logging to better capture authentic human decision-making behavior.

Despite these limitations, this study introduces a novel comparative framework that emphasizes the complementary strengths of humans and interpretable models in phishing detection. By focusing on cognitive reasoning patterns, our work highlights opportunities for leveraging human-machine collaboration to enhance detection strategies, improve interpretability, and inform the design of human-centric AI defenses.

*G. Further Research*

Building on the dual-lens framework established in this study, future research can explore several key directions to enhance the understanding and effectiveness of phishing detection systems.

Firstly studies could integrate larger, more diverse, and real-world phishing datasets, including multilingual samples, to capture a broader range of cues and deception tactics. In addition, experiments could compare human reasoning not only with interpretable models but also with state-of-the-art transformer architectures, analyzed using explainability frameworks such as SHAP or LIME. Finally, incorporating behavioral confidence measures alongside self-reports could yield a more robust understanding of decision certainty in both humans and machines

Second, incorporating more advanced models—such as transformer-based architectures like BERT or multimodal models that consider email structure, headers, and embedded links can help assess trade-offs between performance and interpretability. These models could be paired with explainability tools such as SHAP or LIME to retain transparency.

Third, a longitudinal study capturing human annotation behavior over time would allow us to measure learning effects, fatigue, and adaptation. This can inform the design of targeted training interventions and phishing education campaigns that evolve with user behavior.

Additionally, integrating behavioral signals such as reading time, hesitation, or mouse tracking into the analysis could offer a richer view of human judgment beyond self-reported confidence and emotion. Real-time user studies, possibly in collaboration with email platforms, could validate findings in naturalistic settings.

Finally, developing hybrid systems that blend human-in-the-loop insights with machine predictions, adaptive AI tools that highlight linguistic cues based on common human errors—may yield phishing detection tools that are both robust and aligned with human reasoning patterns. Such tools could be personalized, culturally adaptive, and better suited to support users with varying levels of digital literacy.

# APPENDIX

1. Dear Student,
Our records show that your recent tuition payment failed to process, and your student account is flagged for termination. To avoid losing access to RIT services, please verify your payment information within 24 hours.

Failure to act immediately will result in the suspension of your email, class materials, and transcript access.

Click here to verify your account details

Regards,
RIT IT Services

○ Phishing

○ Not phishing

Confidence Rating Percentage (from 1 to 100 percent)

0   10   20   30   40   50   60   70   80   90   100

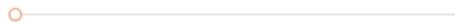

Please note in one sentence which linguistic cues gave away whether the email is phishing or non-phishing?

What emotion did the email invoke in one word?

○ Happiness
○ Anxiety
○ Shame
○ Fear
○ Surprise
○ Angry
○ Other

Fig. 8. Sample of questionnaire